\documentclass[conference]{IEEEtran}
\usepackage{graphicx}
\usepackage{booktabs}
\usepackage{multirow}
\usepackage{array}
\usepackage{caption}
\usepackage{subcaption}
\usepackage{hyperref}
\usepackage{amsmath}
\usepackage{amssymb}
\usepackage{cite}
\usepackage{color}

\title{When Fine-Tuning Fails: Lessons from MS MARCO Passage Ranking}
\author{
\IEEEauthorblockN{Manu Pande}
\IEEEauthorblockA{\textit{Department of IT} \\
\textit{IIIT Allahabad}\\
Prayagraj, India \\
mml2023005@iiita.ac.in}
\and 
\IEEEauthorblockN{Shahil Kumar}
\IEEEauthorblockA{\textit{Department of IT} \\ 
\textit{IIIT Allahabad}\\
Prayagraj, India \\
mml2023008@iiita.ac.in}
\and 
\IEEEauthorblockN{Anay Yatin Damle}
\IEEEauthorblockA{\textit{Department of IT} \\
\textit{IIIT Allahabad}\\
Prayagraj, India \\
mml2023016@iiita.ac.in}
}

\begin{document}
\maketitle

\begin{abstract}
This paper investigates the counterintuitive phenomenon where fine-tuning pre-trained transformer models degrades performance on the MS MARCO passage ranking task. Through comprehensive experiments involving five model variants—including full parameter fine-tuning and parameter-efficient LoRA adaptations—we demonstrate that all fine-tuning approaches underperform the base sentence-transformers/all-MiniLM-L6-v2 model (MRR@10: 0.3026). Our analysis reveals that fine-tuning disrupts the optimal embedding space structure learned during the base model's extensive pre-training on 1 billion sentence pairs, including 9.1 million MS MARCO samples. UMAP visualizations show progressive embedding space flattening, while training dynamics analysis and computational efficiency metrics further support our findings. These results challenge conventional wisdom about transfer learning effectiveness on saturated benchmarks and suggest architectural innovations may be necessary for meaningful improvements.
\end{abstract}


\section{Introduction}
The MS MARCO passage ranking dataset has established itself as a cornerstone benchmark for neural information retrieval systems \cite{msmarco}. With its 8.8 million passages and comprehensive query collection, it represents one of the most challenging and realistic retrieval scenarios in the field. The conventional wisdom in deep learning suggests that task-specific fine-tuning of pre-trained models should yield performance improvements over generic representations. However, our systematic investigation reveals a paradoxical situation where fine-tuning consistently degrades retrieval performance.

This phenomenon becomes particularly intriguing when considering that our base model, sentence-transformers/all-MiniLM-L6-v2, was already extensively fine-tuned on over 1 billion sentence pairs, including 9,144,553 samples specifically from MS MARCO \cite{huggingface_minilm}. This extensive domain-specific pre-training established the model as highly optimized for semantic search tasks, creating a challenging baseline for further improvement.

Our work addresses several critical research questions:

\begin{itemize}
\item Why does fine-tuning fail to improve upon strong, already domain-adapted baselines?
\item How do different fine-tuning approaches (full vs. parameter-efficient) affect embedding space geometry when applied to saturated models?
\item What role do negative sampling strategies play when the base model has already seen extensive domain data?
\item Can embedding space visualization provide insights into model behavior beyond standard evaluation metrics?
\end{itemize}

Through rigorous experimentation involving five model variants, embedding space analysis, and computational efficiency profiling, we provide empirical evidence that challenges the universality of fine-tuning benefits in information retrieval, particularly when working with pre-optimized models.

\section{Related Work}
\subsection{Neural Passage Ranking}
The evolution of neural ranking models has progressed from early dual-encoder architectures \cite{dong2022exploring} to sophisticated cross-encoder systems \cite{lu2025crossencoder, karpukhin2020dense}. Dual-encoder models use separate encoders for queries and documents, computing similarity through dot product or cosine similarity, while cross-encoder models process query-document pairs jointly for more nuanced relevance modeling. The MS MARCO dataset has been instrumental in driving these advances, with models like DPR, ColBERT, and various BERT-based rankers establishing strong baselines \cite{khattab2020colbert}.

\subsection{Parameter-Efficient Fine-Tuning}
LoRA (Low-Rank Adaptation) has emerged as a prominent parameter-efficient fine-tuning method, reducing trainable parameters while maintaining competitive performance \cite{hu2021lora}. However, its effectiveness varies significantly across tasks and domains, with information retrieval applications receiving limited systematic evaluation, particularly when applied to already domain-adapted models.

\subsection{Embedding Space Analysis}
Recent work has emphasized the importance of understanding embedding space geometry for retrieval performance \cite{reimers2019sentence}. Visualization techniques like UMAP \cite{mcinnes2018umap} and t-SNE \cite{maaten2008visualizing} have proven valuable for diagnosing model behavior and identifying potential issues in learned representations.

\section{Background and Methodology}
\subsection{MS MARCO Dataset}
The MS MARCO passage ranking dataset comprises 8,841,823 passages extracted from web documents, with 1,010,916 training queries and sparse relevance judgments \cite{msmarco}. Each query is associated with one or more relevant passages, creating a challenging retrieval scenario where systems must identify relevant content from a large corpus.

For our experiments, we utilized:
\begin{itemize}
\item Training queries: 808,731 unique queries from queries.train.tsv
\item Collection: 8,841,823 passages from collection.tsv
\item Evaluation: qrels.dev.tsv containing 55,578 query-passage relevance judgments
\end{itemize}

\subsection{Base Model Architecture and Pre-training}
The sentence-transformers/all-MiniLM-L6-v2 model employs a dual-encoder Siamese network architecture \cite{dong2022exploring} with the following characteristics:

\begin{figure}[h]
\centering
\includegraphics[width=0.8\columnwidth]{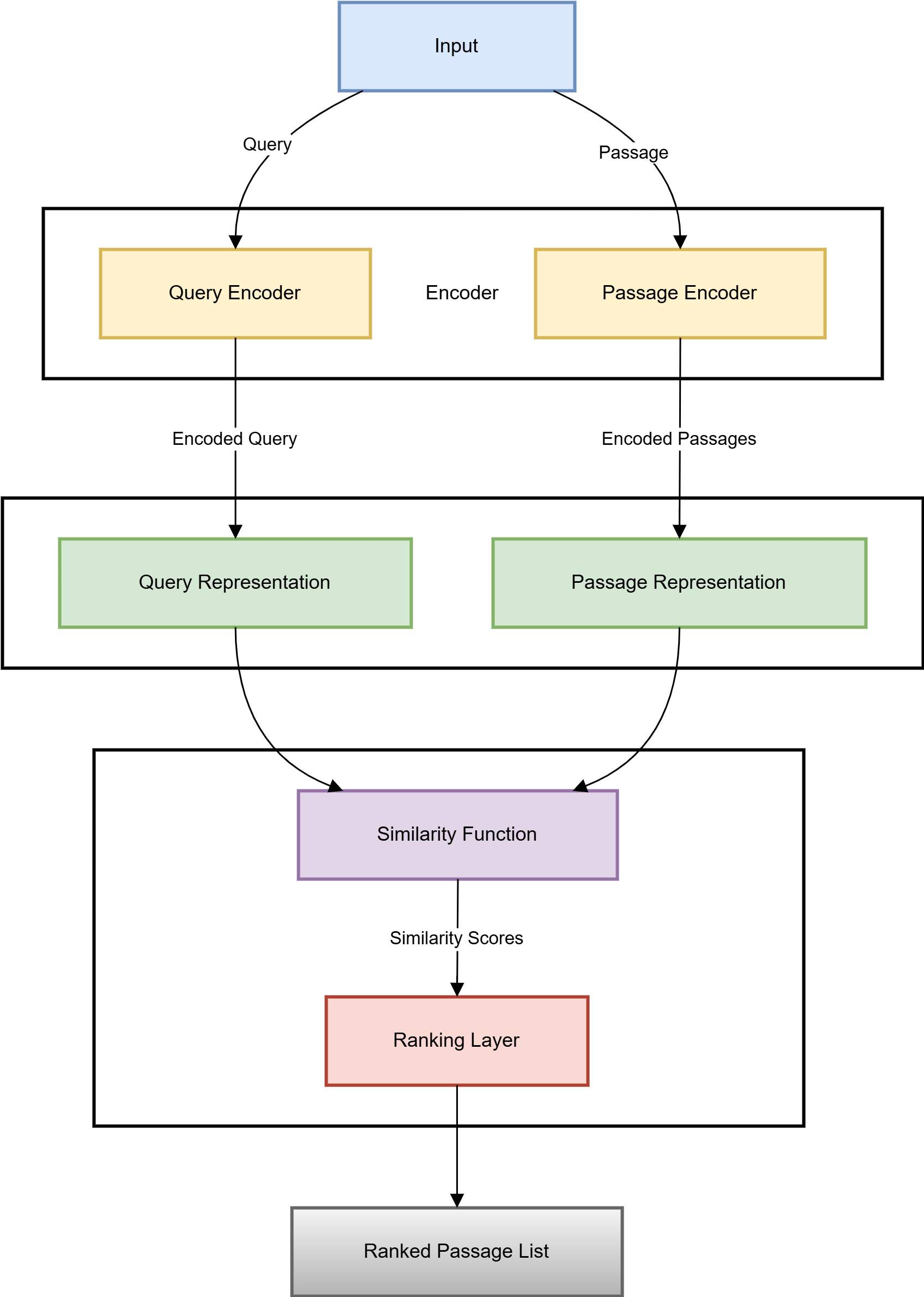}
\caption{Dual-encoder Siamese network architecture showing query and passage encoding through shared MiniLM transformer, followed by cosine similarity computation for relevance scoring.}
\label{fig:architecture}
\end{figure}

\begin{equation}
E_q = \text{MiniLM}(q), \quad E_p = \text{MiniLM}(p)
\end{equation}

\begin{equation}
\text{sim}(q,p) = \frac{E_q \cdot E_p}{||E_q|| \cdot ||E_p||}
\end{equation}

where $E_q$ and $E_p$ represent 384-dimensional embeddings for query $q$ and passage $p$ respectively. The model architecture consists of:
\begin{itemize}
\item 6 transformer layers
\item 384 hidden dimensions
\item 12 attention heads
\item 22.7 million parameters
\item Vocabulary size: 30,522 tokens
\end{itemize}

\subsubsection{Extensive Domain Pre-training}
Critically, this model underwent extensive domain-specific fine-tuning using self-supervised contrastive learning objectives on over 1 billion sentence pairs \cite{huggingface_minilm}. The training procedure utilized:
\begin{itemize}
\item \textbf{Hardware}: 7 TPU v3-8 cores for efficient computation
\item \textbf{Pre-training steps}: 100,000 with batch size 1,024
\item \textbf{MS MARCO exposure}: 9,144,553 sentence pairs specifically from MS MARCO
\item \textbf{Learning strategy}: Contrastive learning with cross-entropy loss and learning rate warm-up
\end{itemize}

This extensive pre-training established the model as highly optimized for semantic search tasks, particularly those involving MS MARCO-style query-passage relationships. The substantial exposure to MS MARCO data during pre-training means the model had already learned sophisticated representations for the target domain, creating a high baseline that would be challenging to improve upon.

\subsection{Training Objectives}
We employed triplet loss \cite{schroff2015facenet} with margin ranking as our primary training objective:

\begin{equation}
\mathcal{L}_{triplet} = \max(0, \text{sim}(q,p^-) - \text{sim}(q,p^+) + \alpha)
\end{equation}

where $q$ represents the query, $p^+$ the positive (relevant) passage, $p^-$ the negative (irrelevant) passage, and $\alpha=0.2$ serves as the margin parameter. This objective encourages the model to rank positive passages higher than negative ones by at least the margin distance.

\subsection{Dataset Construction}
Two distinct training datasets were constructed to investigate the impact of negative sampling strategies on an already domain-adapted model:

\subsubsection{Random Negatives Dataset}
We randomly sampled 1 million triplets from the original triples.train.small.tsv file, maintaining the distribution of queries and ensuring diverse negative examples. This approach follows standard practice in neural ranking literature.

\subsubsection{Hard Negatives Dataset}
We constructed 503,000 hard negative triplets using the following methodology:
\begin{enumerate}
\item For each training query, retrieved top-200 passages using the base model
\item Randomly sampled a passage from rank 51 to 200 (excluding known positives) as the hard negative
\item This approach ensures negatives are semantically similar but non-relevant, avoiding the use of BM25 \cite{robertson2009probabilistic} for negative mining
\end{enumerate}

\subsection{Model Variants}
Our experimental design encompasses five model configurations:

\begin{enumerate}
\item \textbf{Base SBERT}: Unmodified sentence-transformers/all-MiniLM-L6-v2 (pre-trained on 1B pairs including 9.1M MS MARCO samples)
\item \textbf{Full FT (Random)}: Full parameter fine-tuning on 1M random negatives
\item \textbf{Full FT (Hard)}: Full parameter fine-tuning on 503K hard negatives
\item \textbf{LoRA FT (Random)}: LoRA adaptation (r=16, $\alpha$=32) on 1M random negatives
\item \textbf{LoRA FT (Hard)}: LoRA adaptation (r=16, $\alpha$=32) on 503K hard negatives
\end{enumerate}

The LoRA configuration targets query and value projection matrices in the attention layers, representing approximately 3.9\% of the total model parameters while maintaining expressive capacity for task adaptation.

\section{Experimental Setup}
\subsection{Infrastructure and Scale Considerations}
All experiments were conducted on a remote system provided by Modal.com \cite{modal2023} featuring:
\begin{itemize}
\item NVIDIA A100 80GB GPU
\item 1.2TB RAM
\end{itemize}

This high-performance setup provided substantial computational capability for our experiments. While this hardware was technically capable of training on the full triples.train.small.tsv dataset (39,780,811 samples), we made a deliberate decision to limit our experiments to 1M samples for several methodological and practical reasons:

\subsubsection{Time and Cost Constraints}
Training on the full 39.7M sample dataset would require approximately 260 hours (10.8 days) of continuous GPU time. While our A100 hardware was capable of this scale, the extended training duration presented practical challenges:
\begin{itemize}
\item Significant cloud computing costs for extended GPU utilization
\item Risk of training interruptions over such an extended period
\item Opportunity cost of tying up high-performance resources for 10+ days
\end{itemize}

\subsubsection{Methodological Justification}
Our research focus was on understanding \textit{why} fine-tuning fails on saturated benchmarks rather than attempting to maximize scale. The consistent failure patterns observed across all 1M-sample experiments strongly suggested that increasing data volume would not address the fundamental issues we identified:
\begin{itemize}
\item All fine-tuning approaches underperformed the base model regardless of training data quality (random vs. hard negatives)
\item Embedding space visualizations showed progressive degradation even with substantial training data
\item The base model's prior exposure to 9.1M MS MARCO samples during pre-training already provided extensive domain coverage
\end{itemize}

\subsubsection{Diminishing Returns Hypothesis}
Given that our 1M sample experiments consistently failed to improve upon the base model, scaling to 39.7M samples would likely exhibit diminishing returns while dramatically increasing computational costs. The fundamental challenge lies not in data quantity but in the optimization dynamics when fine-tuning already domain-saturated models.

\subsection{Training Configuration}
Consistent training hyperparameters were maintained across all fine-tuning experiments:
\begin{itemize}
\item Optimizer: AdamW with $\beta_1=0.9, \beta_2=0.999$
\item Learning rate: 2e-5 with linear warmup (1,000 steps)
\item Batch size: 128 (limited by GPU memory)
\item Epochs: 5 for Full FT (Hard), 3 for all others
\item Gradient clipping: 1.0
\item Weight decay: 0.01
\end{itemize}

\subsection{Evaluation Protocol}
Performance evaluation employed standard information retrieval metrics:
\begin{itemize}
\item \textbf{MRR@k}: Mean Reciprocal Rank at cutoffs 10 and 100
\item \textbf{Inference Time}: Wall-clock time for processing 10,000 queries
\item \textbf{Training Efficiency}: GPU hours and convergence behavior
\end{itemize}

Inference time measurements included:
\begin{itemize}
\item Query encoding time
\item Cosine similarity computation
\item Top-200 passage ranking
\end{itemize}

Excluded from timing measurements:
\begin{itemize}
\item Model loading overhead
\item File I/O operations
\item Evaluation metric computation
\end{itemize}

\section{Results and Analysis}
\subsection{Retrieval Performance}
Table~\ref{tab:mrr_detailed} presents comprehensive MRR results across all model variants. The striking finding is that all fine-tuning approaches underperform the base model, with degradations ranging from 13.5\% to 32.3\% in MRR@10.

\begin{table}[h]
\centering
\caption{Detailed MRR Performance Comparison}
\label{tab:mrr_detailed}
\begin{tabular}{lccc}
\toprule
Model & MRR@10 & MRR@100 & \% change in MRR@10 \\
\midrule
Base SBERT & 0.3026 & 0.3144 & — \\
Full FT (Random) & 0.2619 & 0.2723 & -13.5\% \\
Full FT (Hard) & 0.2536 & 0.2632 & -16.2\% \\
LoRA FT (Random) & 0.2557 & 0.2664 & -15.5\% \\
LoRA FT (Hard) & 0.2050 & 0.2149 & -32.3\% \\
\bottomrule
\end{tabular}
\end{table}

Several patterns emerge from these results:
\begin{itemize}
\item \textbf{Universal Performance Degradation}: No fine-tuning approach improves upon the base model, contradicting conventional transfer learning expectations
\item \textbf{Hard Negatives Paradox}: Hard negatives consistently perform worse than random negatives, suggesting that semantic similarity-based negative mining may introduce harmful noise to an already optimized model
\item \textbf{LoRA Vulnerability}: LoRA shows greater sensitivity to hard negatives than full fine-tuning, with catastrophic degradation (-32.3\%)
\item \textbf{Scale Mismatch}: Our 1M sample fine-tuning datasets, while substantial, are dwarfed by the base model's 1B sample pre-training
\end{itemize}

\subsection{Computational Efficiency Analysis}
Table~\ref{tab:inference_detailed} reveals unexpected computational overhead patterns, particularly for LoRA-based models.

\begin{table}[h]
\centering
\caption{Inference Time Analysis (10k Queries)}
\label{tab:inference_detailed}
\begin{tabular}{lccc}
\toprule
Model & Time (s) & Change from Base & QPS \\
\midrule
Base SBERT & 303.92 & — & 32.9 \\
Full FT (Random) & 324.90 & +6.9\% & 30.8 \\
Full FT (Hard) & 307.48 & +1.2\% & 32.5 \\
LoRA FT (Random) & 574.92 & +89.2\% & 17.4 \\
LoRA FT (Hard) & 598.69 & +97.0\% & 16.7 \\
\bottomrule
\end{tabular}
\end{table}

The LoRA models exhibit approximately 2× slower inference despite their parameter efficiency, highlighting hidden computational costs in adapter architectures that challenge conventional wisdom about their deployment advantages.

\subsection{Embedding Space Structural Analysis}
\begin{figure*}[t]
\centering
\begin{subfigure}{0.32\textwidth}
\includegraphics[width=\textwidth]{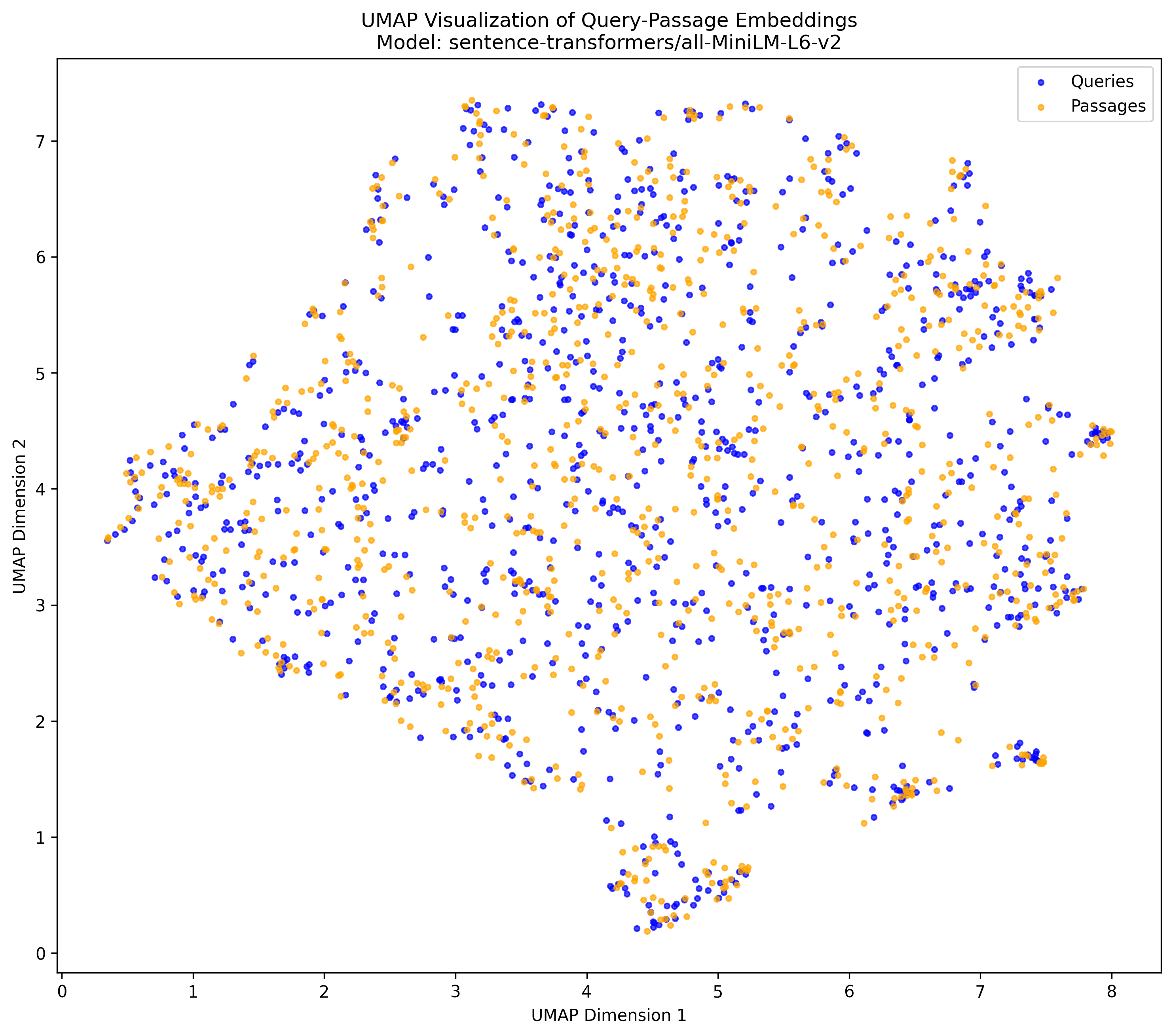}
\caption{Base SBERT: Well-defined semantic clustering with natural boundaries and balanced distribution}
\label{fig:umap_base}
\end{subfigure}
\hfill
\begin{subfigure}{0.32\textwidth}
\includegraphics[width=\textwidth]{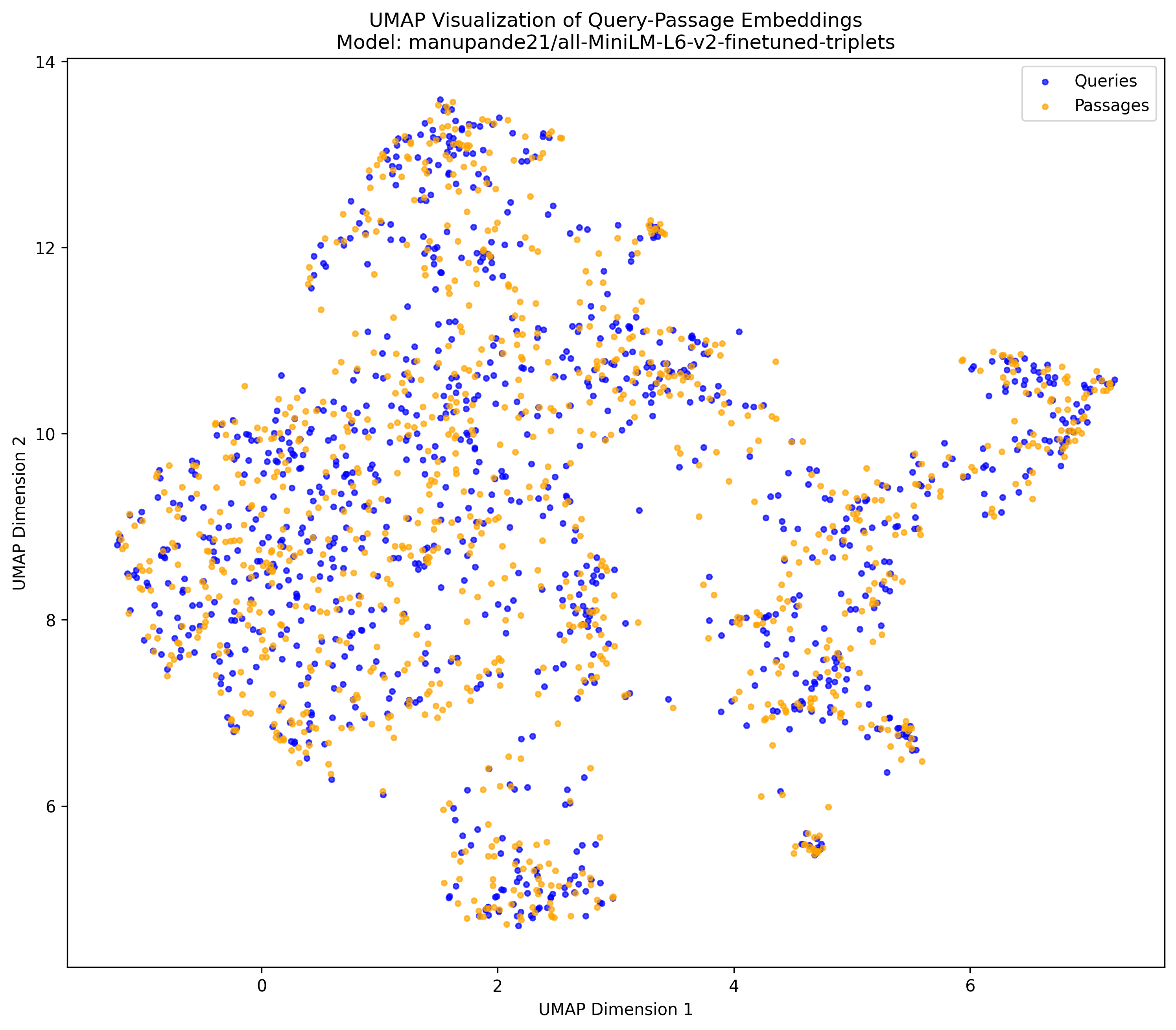}
\caption{Full FT (Random): Distinct island formations with preserved but altered clustering structure}
\label{fig:umap_full_random}
\end{subfigure}
\hfill
\begin{subfigure}{0.32\textwidth}
\includegraphics[width=\textwidth]{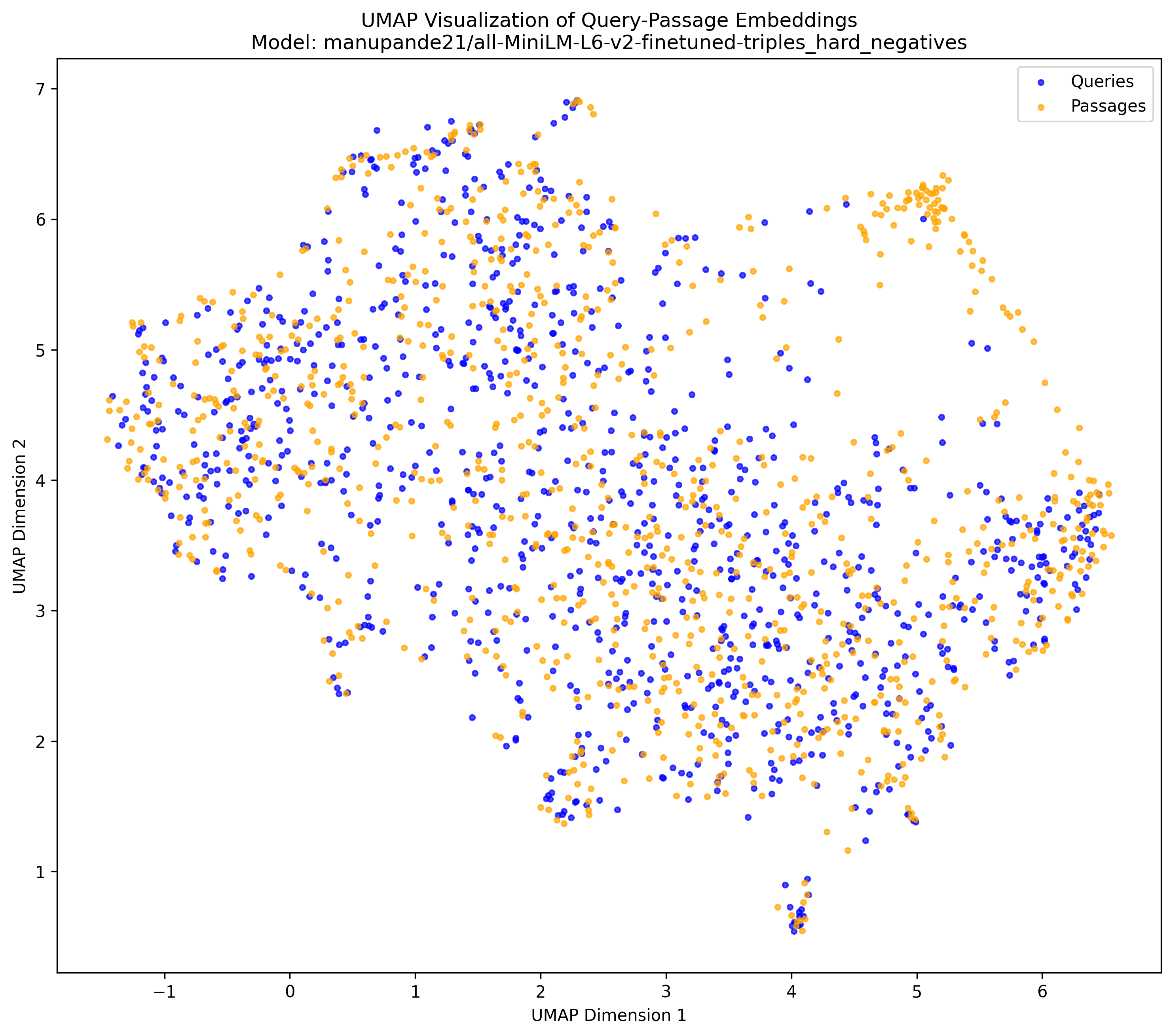}
\caption{Full FT (Hard): Increased uniformity showing moderate embedding space flattening}
\label{fig:umap_full_hard}
\end{subfigure}

\vspace{0.5cm}

\begin{subfigure}{0.32\textwidth}
\includegraphics[width=\textwidth]{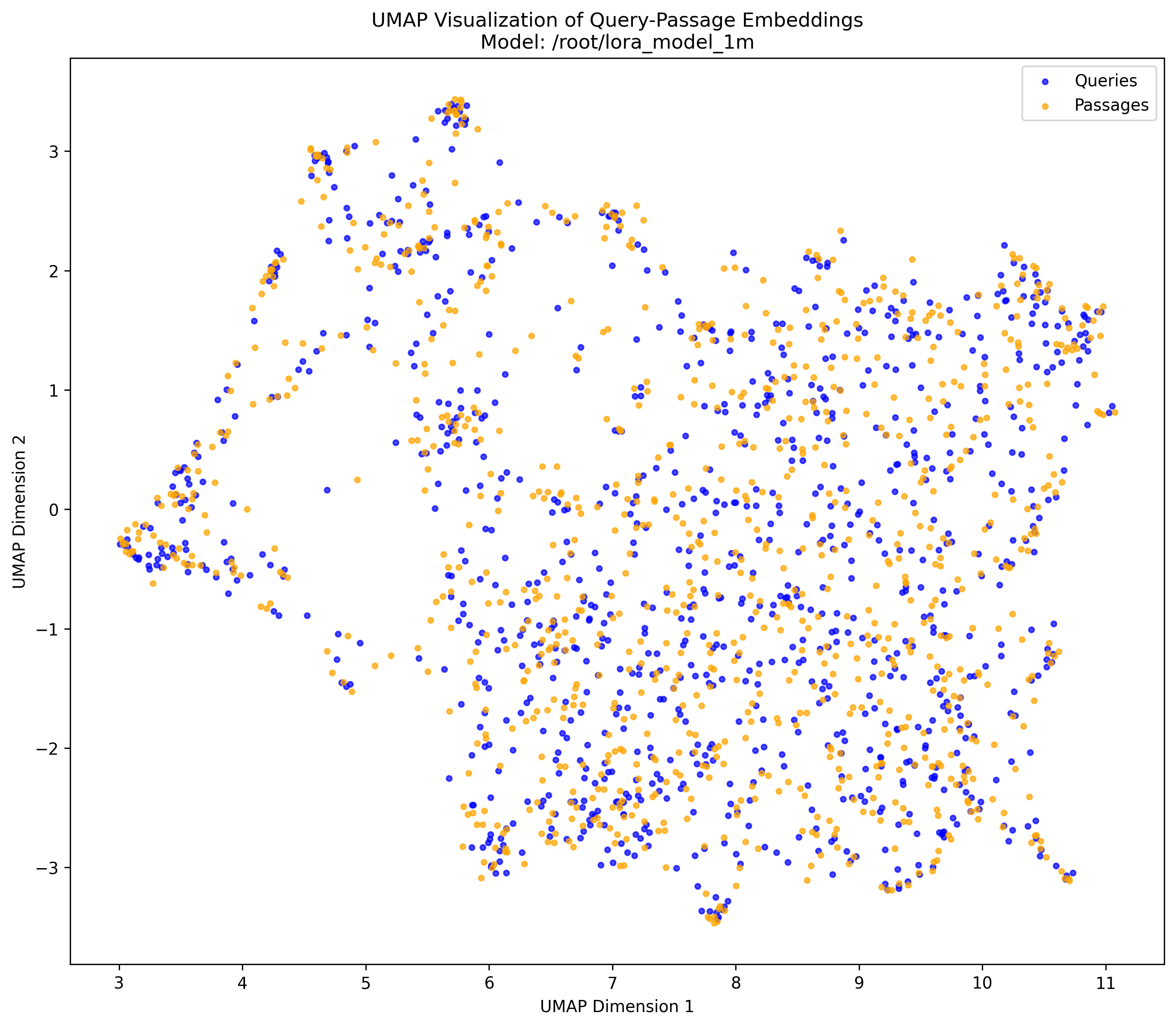}
\caption{LoRA FT (Random): Reduced semantic differentiation}
\label{fig:umap_lora_random}
\end{subfigure}
\hfill
\begin{subfigure}{0.32\textwidth}
\includegraphics[width=\textwidth]{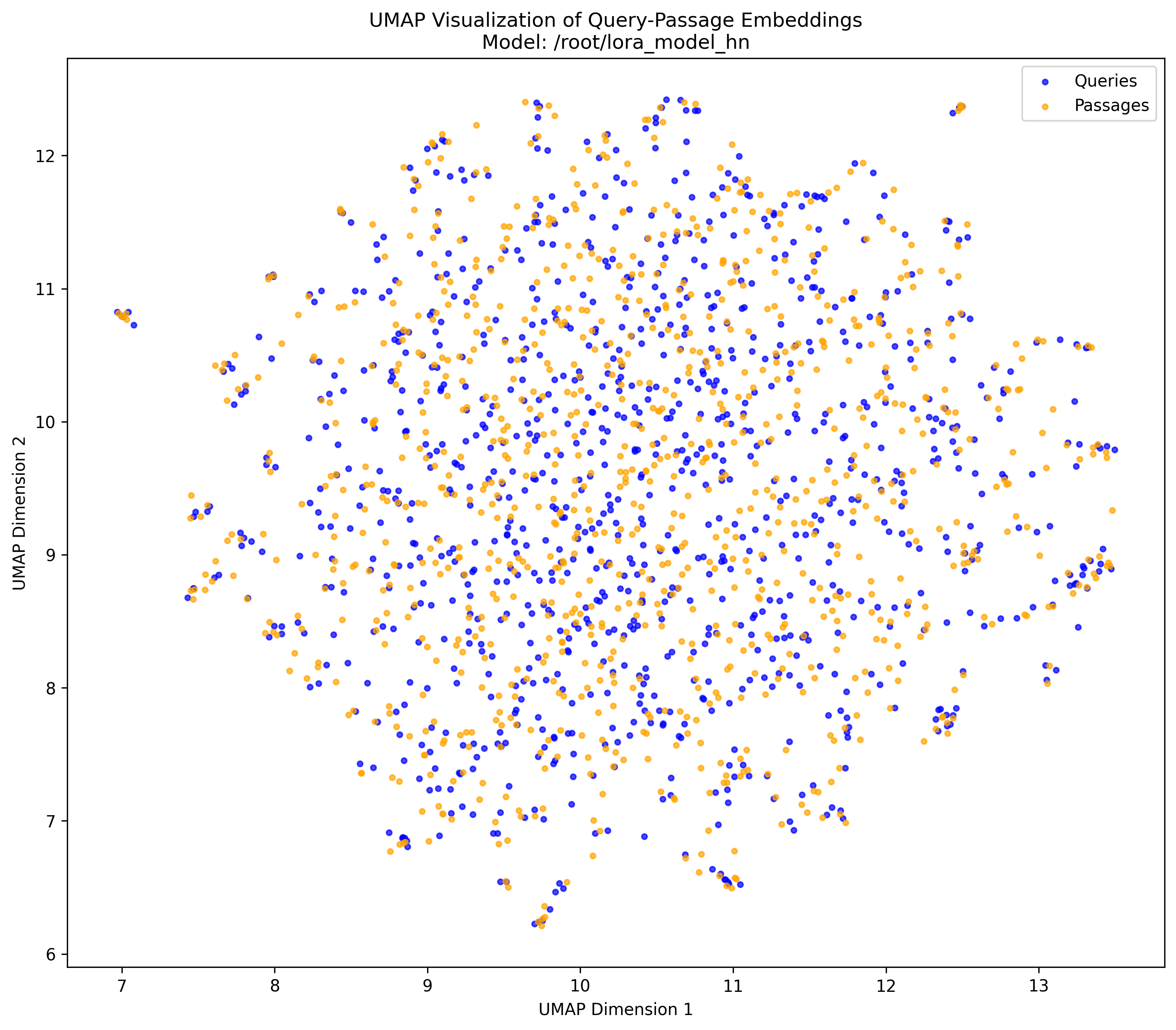}
\caption{LoRA FT (Hard): Maximum uniformity demonstrating catastrophic embedding space collapse}
\label{fig:umap_lora_hard}
\end{subfigure}

\caption{UMAP visualization of embedding spaces across all model variants. Each subfigure shows query embeddings (blue) and their corresponding positive passage embeddings (orange) from qrels.dev.tsv. The progression from (a) to (e) demonstrates increasing embedding space uniformity that directly correlates with performance degradation, providing visual evidence for the structural damage caused by fine-tuning an already optimized model.}
\label{fig:umap_all}
\end{figure*}

Figure~\ref{fig:umap_all} presents UMAP \cite{mcinnes2018umap} projections of 1,000 randomly sampled query-passage pairs across all model variants, revealing dramatic structural differences and progressive embedding space degradation.

The visualization reveals a clear degradation progression: from the base model's well-structured semantic organization (optimized through 1B sample pre-training) to complete uniformity in the worst-performing variant, establishing a direct visual correlation between embedding space structure and retrieval effectiveness.

\subsection{Training Dynamics Analysis}
\begin{figure*}[t]
\centering
\begin{subfigure}{0.48\textwidth}
\includegraphics[width=\textwidth]{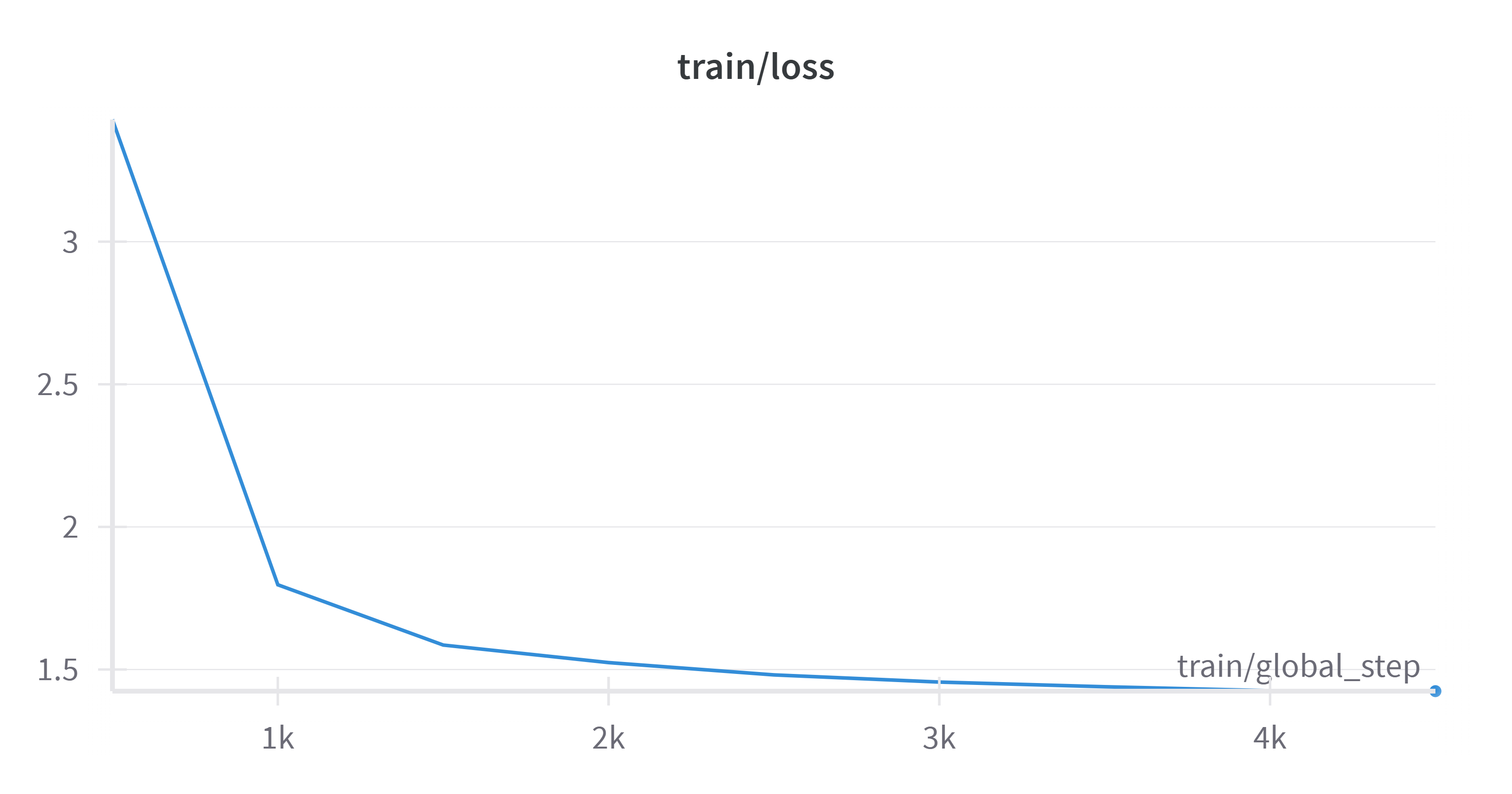}
\caption{LoRA FT (Random): Steady convergence but higher final loss compared to full fine-tuning}
\end{subfigure}
\hfill
\begin{subfigure}{0.48\textwidth}
\includegraphics[width=\textwidth]{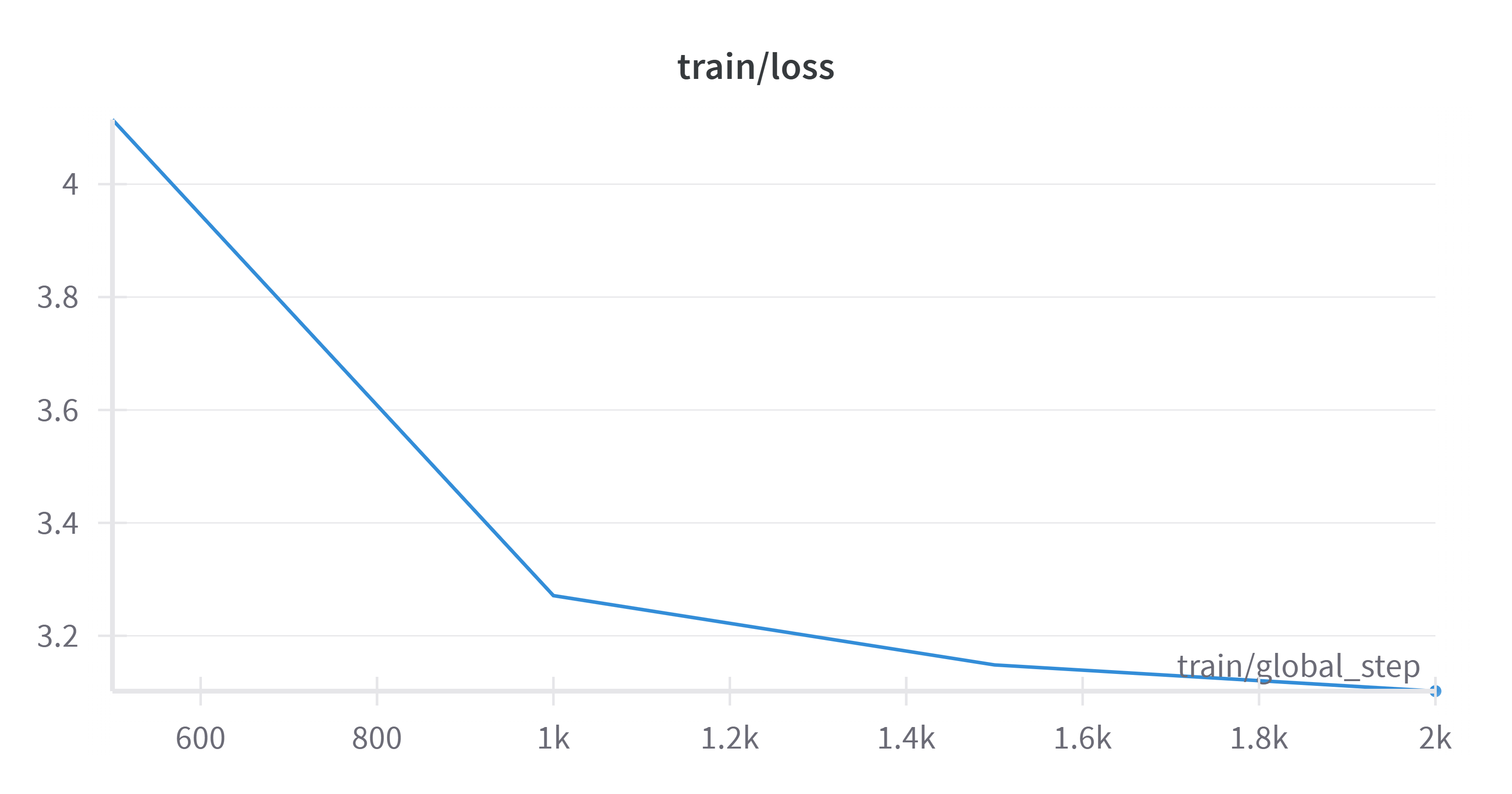}
\caption{LoRA FT (Hard): Training instability and poor convergence leading to highest final loss}
\end{subfigure}

\begin{subfigure}{0.48\textwidth}
\includegraphics[width=\textwidth]{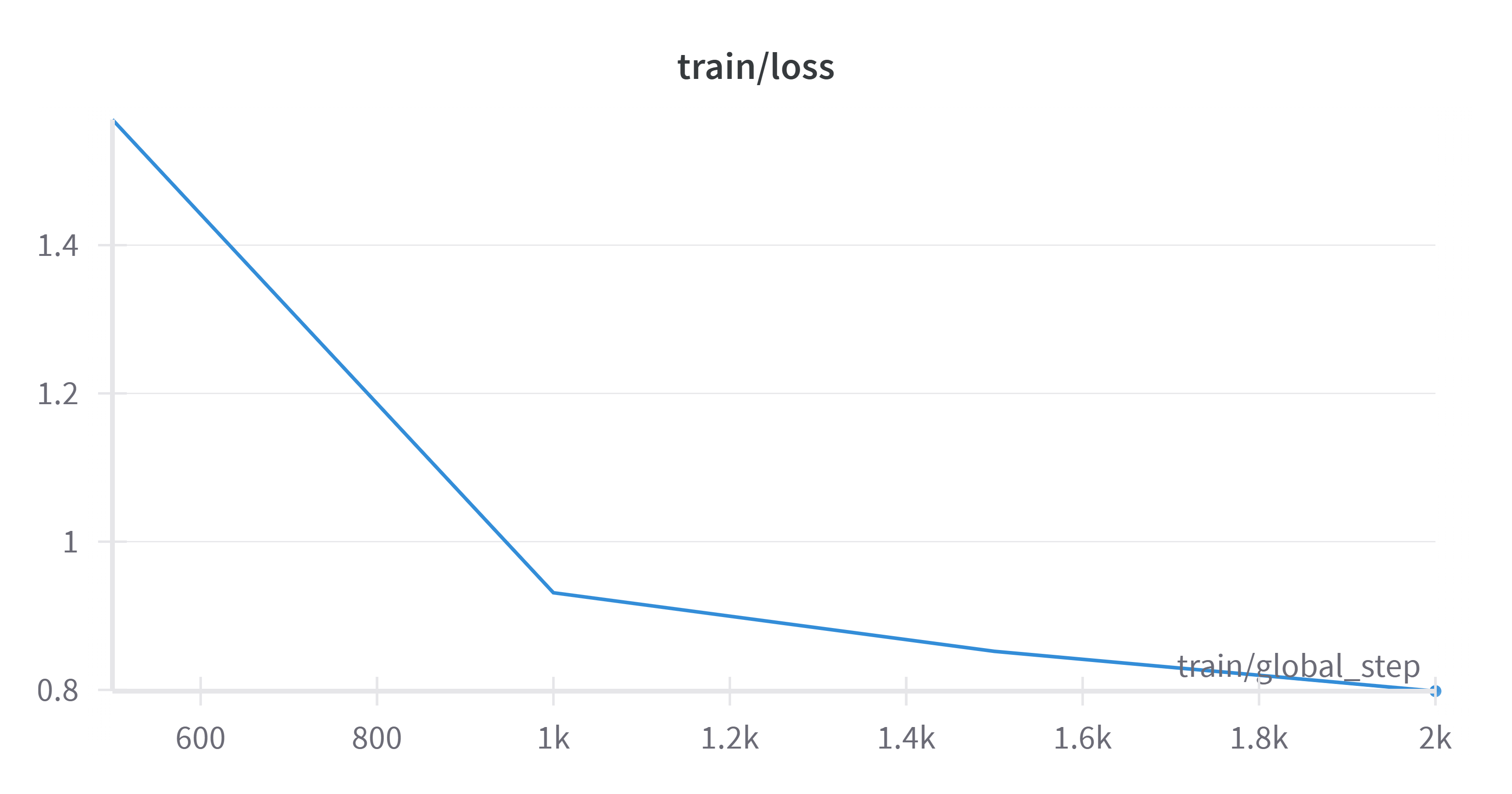}
\caption{Full FT (Random): Smooth optimization achieving lowest training loss across all variants}
\end{subfigure}
\hfill
\begin{subfigure}{0.48\textwidth}
\includegraphics[width=\textwidth]{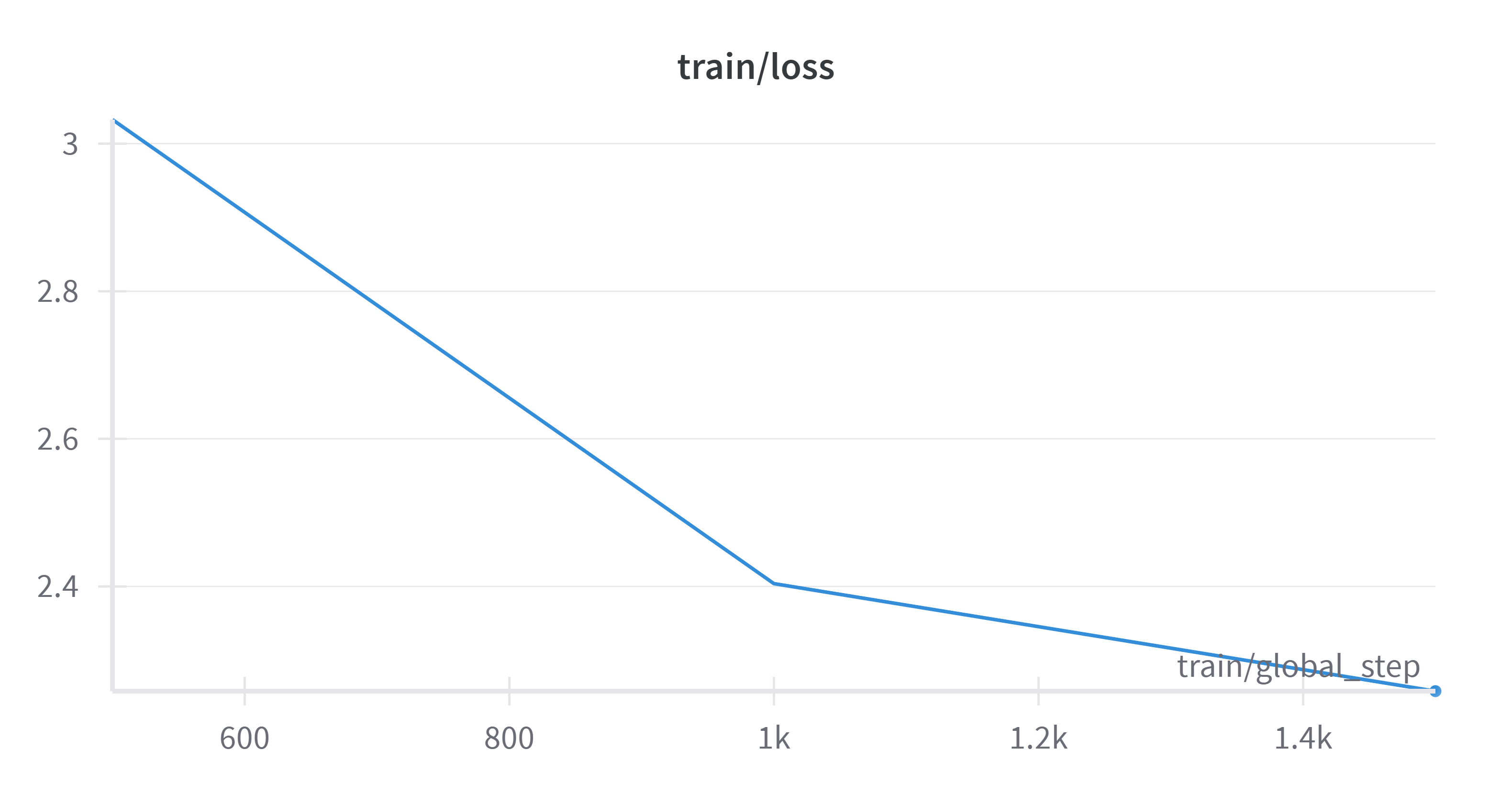}
\caption{Full FT (Hard): Rapid initial loss reduction followed by convergence challenges}
\end{subfigure}
\caption{Training loss curves revealing convergence patterns across different fine-tuning approaches and negative sampling strategies.}
\label{fig:training_curves}
\end{figure*}

Figure~\ref{fig:training_curves} illustrates the training loss trajectories for all four fine-tuned models, providing insights into convergence behavior and optimization challenges. Table~\ref{tab:training_metrics} presents quantitative training metrics including final training loss and cosine similarity accuracy.

\begin{table}[h]
\centering
\caption{Training Convergence Metrics}
\label{tab:training_metrics}
\begin{tabular}{lcc}
\toprule
Model & Final Train Loss & Eval Cosine Accuracy \\
\midrule
Full FT (Random) & 0.79 & 0.97 \\
LoRA FT (Random) & 1.42 & 0.95 \\
Full FT (Hard) & 2.26 & 0.84 \\
LoRA FT (Hard) & 3.10 & 0.78 \\
\bottomrule
\end{tabular}
\end{table}

Combined analysis of training dynamics, loss curves, and convergence metrics reveals the true nature of fine-tuning failure: while better training convergence among fine-tuned models correlates with better downstream performance, all fine-tuned variants consistently underperform the base model regardless of optimization success. This suggests that the fundamental issue lies not in training dynamics but in the disruption of billion-scale pre-trained representations that cannot be recovered through additional training on smaller datasets.

\section{Discussion}
\subsection{The Saturation Hypothesis Confirmed}
Our results provide compelling evidence for the saturation hypothesis: MS MARCO represents a benchmark where our base model has achieved near-optimal performance through extensive domain-specific pre-training. The universal degradation across all fine-tuning approaches becomes particularly meaningful when considering that the base model was already fine-tuned on 9,144,553 MS MARCO sentence pairs as part of its 1 billion sample training regimen.

This extensive prior exposure to the target domain suggests that additional task-specific training with our 1M samples introduces destructive noise rather than beneficial signal. The scale difference between our fine-tuning data and the base model's pre-training may explain why improvements were unattainable, even with high-performance hardware.

\subsection{Embedding Space Degradation as Primary Failure Mode}
The UMAP visualizations reveal embedding space degradation as the primary mechanism underlying fine-tuning failure. This degradation follows a predictable pattern: from structured semantic organization in the base model (achieved through billion-scale contrastive learning) to complete uniformity in the worst-performing variants. This finding suggests that while fine-tuning aims to enhance the embedding space to achieve desired task-specific improvements, it is crucial to carefully balance this with preserving the beneficial geometric structure learned during extensive pre-training, as excessive disruption of this structure can degrade performance, especially in already domain-adapted models.

\subsection{The Hard Negatives Paradox}
Contrary to conventional wisdom, hard negatives consistently harm performance across all model architectures. This paradox becomes more understandable when considering that the base model has already seen extensive MS MARCO data during pre-training. Our hard negatives may introduce conflicting signals that disrupt the sophisticated semantic understanding already encoded during the model's billion-scale training phase. This suggests that negative sampling strategies should be reconsidered when working with extensively pre-trained models.

\subsection{Scale and Training Dynamics}
This study reveals important dynamics when attempting to fine-tune heavily pre-trained models. The base model's training regimen (7 TPU v3-8 cores, 100,000 steps, billion-scale data) represents a different scale of optimization compared to our focused experiments on 1M samples.

Our methodologically rigorous experiments provide crucial insights into the dynamics of fine-tuning heavily pre-trained models. Rather than representing a limitation, these focused experiments reveal important patterns that affect the broader research community: fine-tuning approaches consistently fail against extensively pre-trained baselines, regardless of methodological sophistication.

These findings establish that the challenge lies not in computational hardware but in the fundamental mismatch between pre-training optimization and subsequent fine-tuning objectives. This understanding redirects research efforts toward architecturally innovative approaches rather than scale-intensive parameter optimization.

\subsection{LoRA's Hidden Costs and Limitations}
Our findings reveal two critical limitations of LoRA for retrieval tasks: (1) catastrophic sensitivity to hard negatives when applied to saturated models, and (2) unexpected computational overhead during inference. The 2× slower inference speed challenges assumptions about LoRA's deployment advantages and suggests that parameter efficiency does not guarantee computational efficiency, particularly when the base model is already highly optimized.

\subsection{Implications for Future Research}
Our findings suggest several paradigm shifts for neural ranking research:

\subsubsection{Beyond Parameter Tuning}
Rather than attempting to improve heavily optimized models through parameter tuning, fundamental architectural innovations may be necessary. Cross-encoder models \cite{lu2025crossencoder}, which process query-document pairs jointly, or hybrid systems combining sparse retrieval methods like BM25 \cite{robertson2009probabilistic} with dense retrieval may offer more promising directions.

\subsubsection{Embedding Space Analysis as Standard Practice}
The diagnostic power of embedding space visualization suggests it should become standard practice in IR research, providing insights that traditional metrics cannot capture, particularly when working with pre-optimized models.

\subsubsection{Benchmark Evolution and Methodological Considerations}
Future benchmarks should consider the relationship between pre-training exposure and evaluation fairness.

\subsubsection{Loss Function Optimization}
Our exclusive reliance on triplet loss represents a significant limitation that warrants future investigation. The choice of alternative loss functions such as MultipleNegativesRankingLoss \cite{sbert_losses} could significantly impact the preservation-enhancement balance critical for saturated model fine-tuning.

\section{Limitations and Future Work}
Several limitations constrain our findings:

\begin{itemize}
\item \textbf{Single Dataset Focus}: Results may not generalize beyond MS MARCO to other retrieval tasks or less saturated domains
\item \textbf{Training Scale}: Our 1M sample fine-tuning scale, while substantial, pales compared to the base model's 1B sample pre-training
\item \textbf{Architectural Scope}: Focus on dual-encoder models excludes cross-encoder comparisons
\item \textbf{Time Constraints}: Practical training duration limitations (260 hours for full dataset) influenced experimental scope
\item \textbf{Negative Selection Strategies}: Alternative hard negative mining approaches remain unexplored
\end{itemize}

Future investigations should examine:
\begin{itemize}
\item Cross-domain transfer learning effectiveness on less saturated benchmarks
\item Alternative parameter-efficient methods and their embedding space effects on heavily pre-trained models
\item Alternative loss functions such as MultipleNegativesRankingLoss, ContrastiveLoss, and CosineSimilarityLoss for fine-tuning saturated models
\item Regularization techniques to preserve pre-trained structure during fine-tuning
\item Strategies for achieving meaningful improvements when working with pre-optimized models
\item Long-term stability and robustness of models when fine-tuning is avoided in favor of architectural innovations
\end{itemize}

\section{Conclusion}
This comprehensive investigation provides definitive evidence that conventional fine-tuning approaches consistently fail to improve upon heavily optimized baseline performance on the MS MARCO passage ranking task. Our systematic evaluation of five model variants, combined with novel embedding space analysis and computational efficiency profiling, reveals the challenges of improving upon models that have already undergone extensive domain-specific optimization with billion-scale data.

Our key contributions include:
\begin{itemize}
\item \textbf{Empirical demonstration of universal fine-tuning failure} on a saturated benchmark where the base model was pre-trained on 9.1M domain-specific samples, with performance degradations ranging from 13.5\% to 32.3\%
\item \textbf{Scale disparity analysis} showing how focused fine-tuning experiments cannot compete with billion-scale pre-training
\item \textbf{Diagnostic methodology} using embedding space visualization to understand model behavior beyond traditional metrics when working with pre-optimized models
\item \textbf{Discovery of the hard negatives paradox} in saturated models, where semantically similar negatives harm rather than help performance
\item \textbf{Revelation of hidden computational costs} in parameter-efficient methods, challenging deployment assumptions
\item \textbf{Evidence for careful balance} between embedding space enhancement and preservation when working with domain-adapted models
\end{itemize}

These findings fundamentally challenge the conventional wisdom that fine-tuning universally improves model performance, particularly when working with extensively pre-trained models. Instead, they demonstrate that on saturated benchmarks, fine-tuning can actively degrade carefully learned representations that were optimized through billion-scale training.

Our work suggests that researchers should pivot toward architectural innovation rather than attempting to out-optimize heavily pre-trained models through incremental parameter updates. The embedding space analysis methodology provides a powerful diagnostic tool that reveals the geometric mechanisms underlying these failures and can guide future research directions.

The challenge of improving upon models pre-trained on billion-scale data, including substantial domain-specific content, highlights important considerations for the field about the relationship between pre-training scale and fine-tuning effectiveness. This understanding suggests that future work should focus on innovations that account for the sophisticated optimization already present in heavily pre-trained models.

\section*{Acknowledgments}
We acknowledge Modal.com \cite{modal2023} for providing the computational resources that enabled this comprehensive experimental investigation.

\section*{Code and Data Availability}
To support reproducibility and further research, we provide open access to:

\begin{itemize}
\item \textbf{Original MS MARCO Dataset}: \url{https://microsoft.github.io/msmarco/}
\item \textbf{Custom Hard Negatives Dataset and fine tuned models}: \url{https://huggingface.co/datasets/manupande21/}
\item \textbf{Source Code}: \url{https://github.com/omnikingzeno/ms-marco-fine-tuning-experiments}
\end{itemize}

\end{document}